\documentclass{article}

\PassOptionsToPackage{numbers, square, sort&compress}{natbib}

\usepackage[preprint]{nips_2018}

\usepackage[utf8]{inputenc} 
\DeclareUnicodeCharacter{2212}{-}
\usepackage[T1]{fontenc}    
\usepackage{hyperref}       
\usepackage{url}            
\usepackage{booktabs}       
\usepackage{amsfonts}       
\usepackage{nicefrac}       
\usepackage{microtype}      

\usepackage{multirow}
\usepackage{subcaption}

\usepackage{amsmath}
\usepackage{graphicx}
\graphicspath{{img/}}

\RequirePackage{scrlfile}
\PreventPackageFromLoading{subfig}
\usepackage{svg}

\usepackage{authblk}

\title{Stochastic Adaptive Neural Architecture Search for Keyword Spotting}

\author[ \dag]{Tom V\'eniat}
\author[ \dag]{Olivier Schwander}
\author[ \dag, *]{Ludovic Denoyer}
\affil[ \dag]{Sorbonne Universit\'e, LIP6, F-75005, Paris, France}
\affil[*]{Facebook AI Research}
\affil[ ]{\texttt{\{tom.veniat, olivier.schwander\}@lip6.fr}}
\affil[ ]{\texttt{denoyer@fb.com}}

\begin{document}

\maketitle

\begin{abstract}


The problem of keyword spotting \textit{i.e.} identifying keywords in a real-time audio stream is mainly solved by applying a neural network over successive sliding windows. Due to the difficulty of the task, baseline models are usually large, resulting in a high computational cost and energy consumption level. We propose a new method called SANAS (Stochastic Adaptive Neural Architecture Search) which is able to adapt the architecture of the neural network on-the-fly at inference time such that small architectures will be used when the stream is easy to process (silence, low noise, ...) and bigger networks will be used when the task becomes more difficult. We show that this adaptive model can be learned end-to-end by optimizing a trade-off between the prediction performance and the average computational cost per unit of time. Experiments on the Speech Commands dataset \cite{DBLP:Warden_speech_commands} show that this approach leads to a high recognition level while being much faster (and/or energy saving) than classical approaches where the network architecture is static.
\end{abstract}

\section{Introduction and Related Work}
\label{sec:intro}

Neural Networks (NN) are known to obtain very high recognition rates on a large variety of tasks, and especially over signal-based problems like speech recognition \cite{DBLP:journals/corr/AmodeiABCCCCCCD15}, image classification \cite{DBLP:journals/corr/HeZRS15,DBLP:journals/corr/abs-1802-01548}, etc. However these models are usually composed of millions of parameters involved in millions of operations and have high computational and energy costs at prediction time. There is thus a need to increase their processing speed and reduce their energy footprint. 

From the NN point of view, this problem is often viewed as a problem of network architecture discovery and solved with Neural Architecture Search (NAS) methods in which the search is guided by a trade-off between prediction quality and prediction cost \cite{Veniat_2018_CVPR,DBLP:journals/corr/HuangW17aa,DBLP:journals/corr/abs-1711-06798}. Recent approaches involve for instance Genetic Algorithms \cite{DBLP:journals/corr/RealMSSSLK17, DBLP:journals/corr/abs-1802-01548} or Reinforcement Learning \cite{DBLP:journals/corr/ZophL16, DBLP:journals/corr/ZophVSL17}. While these models often rely on expensive training procedures where multiple architectures are trained, some recent works have proposed to simultaneously discover the architecture of the network while learning its parameters \cite{Veniat_2018_CVPR, DBLP:journals/corr/HuangW17aa} resulting in models that are fast both at training and at inference time. But in all these works, the discovered architecture is static  \textit{i.e.} the same NN being re-used for all the predictions. 

 When dealing with streams of information, reducing the computational and energy costs is of crucial importance. 
 For instance, let us consider the keyword spotting\footnote{See Section \ref{sec:experiments} for a formal description.} problem which is the focus of this paper. It consists in detecting keywords in an audio stream and is particularly relevant for virtual assistants which must continuously listen to their environments to spot user interaction requests. This requires detecting when a word is pronounced, which word has been pronounced and able to run quickly on resource-limited devices. Some recent works \cite{DBLP:conf/interspeech/SainathP15, DBLP:journals/corr/ArikKCHGFPC17, DBLP:conf/icassp/TangL18} proposed to use convolutional neural networks (CNN) in this streaming context, applying a particular model to successive sliding windows \cite{DBLP:conf/interspeech/SainathP15, DBLP:conf/icassp/TangL18} or combining CNNs with recurrent neural networks (RNN) to keep track  of the context \cite{DBLP:journals/corr/ArikKCHGFPC17}. In such cases, the resulting system spends the same amount of time to process each audio frame, irrespective of the content of the frame or its context. 

Our conjecture is that, when dealing with streams of information, a model able to adapt its architecture to the difficulty of the prediction problem at each timestep -- i.e. a small architecture being used when the prediction is easy, and a larger architecture being used when the prediction is more difficult -- would be more efficient than a static model, particularly in terms of computation or energy consumption. 
To achieve this goal, we propose the SANAS algorithm (Section \ref{sec:sanas}): it is, as far as we know, the first architecture search method producing a system which dynamically adapts the architecture of a neural network during prediction at each timestep and which is learned end-to-end by minimizing a trade-off between computation cost and prediction loss.
After learning, our method can process audio streams at a higher speed than classical static methods while keeping a high recognition rate, spending more prediction time on complex signal windows and less time on easier ones (see Section \ref{sec:experiments}).

\begin{figure}[t]
    \centering
    \includegraphics[width=\linewidth]{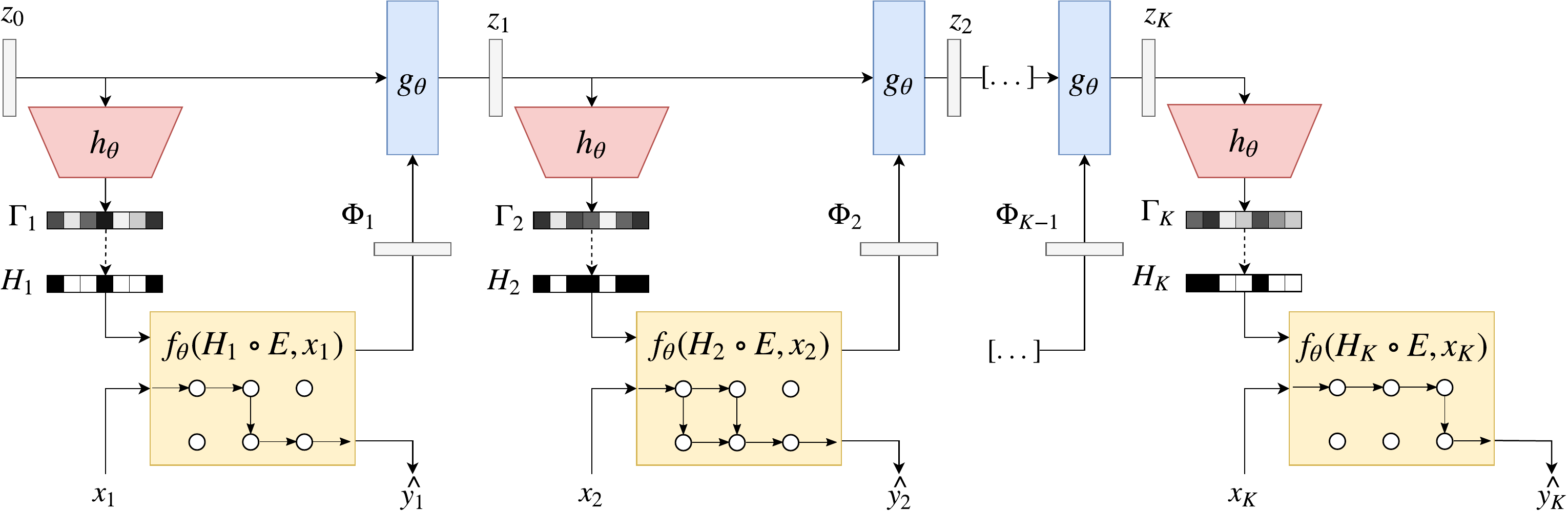}
    \caption{\textbf{SANAS Architecture}. At timestep $t$, the distribution $\Gamma_t$ is generated from the previous hidden state, $\Gamma_t= h(z_t,\theta)$. A discrete architecture $H_t$ is then sampled from $\Gamma_t$ and evaluated over the input $x_t$. This evaluation gives both a feature vector $\Phi(x_t,\theta, E \circ H_t)$ to compute the next hidden state, and the prediction of the model $\hat{y_t}$ using $f(z_t,x_t,\theta, E \circ H_t)$. Dashed edges represent sampling operations. At inference, the architecture which has the highest probability is chosen at each timestep.}
    \label{fig:sanas_arch}
\end{figure}
\section{Adaptive Neural Architecture Search}
\label{sec:method}

\subsection{Problem Definition}\label{sec:classic}

We consider the generic problem of stream labeling where, at each timestep, the system receives a datapoint denoted $x_t$ and produces an output label $y_t$. In the case of audio streams, $x_t$ is usually a time-frequency feature map, and $y_t$ is the presence or absence of a given keyword. In classical approaches, the  output label $y_t$ is  predicted using a neural network whose architecture\footnote{ a precise definition of the notion of architecture is given further.} is denoted $\mathcal{A}$ and whose parameters are $\theta$. We consider in this paper the recurrent modeling scheme where the context $x_1,y_1,.....,x_{t-1},y_{t-1}$ is encoded using a latent representation $z_t$, such that the prediction at time $t$ is made computing $f(z_t,x_t,\theta,\mathcal{A})$, $z_t$ being updated at each timestep such that $z_{t+1}=g(z_t, x_t, \theta,\mathcal{A})$ - note that $g$ and $f$ can share some common computations.

For a particular architecture $\mathcal{A}$, the parameters are learned over a training set of labeled sequences $\{(x^i,y^i)\}_{i \in [1,N]}$, $N$ being the size of the training set, by solving:
\[
\theta^* = \arg \min_\theta \frac{1}{N} \sum\limits_{i=1}^{N} \big[ \sum\limits_{t=1}^{\#x^i} \Delta(f(z_t,x_t,\theta,\mathcal{A}),y_t) \big]
\]
where $\#x^i$ is the length of sequence $x^i$, and $\Delta$ a differentiable loss function. At inference, given a new stream $x$, each label $\hat{y_t}$ is predicted by computing $f(x_1,\hat{y_1},.....,\hat{y_{t-1}},x_t,\theta^ *,\mathcal{A})$, where $\hat{y_1} \ldots \hat{y_{t-1}}$ are the predictions of the model at previous timesteps. In that case, the computation cost of each prediction step solely depends on the architecture and is denoted $C(\mathcal{A})$. 

\begin{figure}[t!]
    \centering
    \includegraphics[width=1\linewidth]{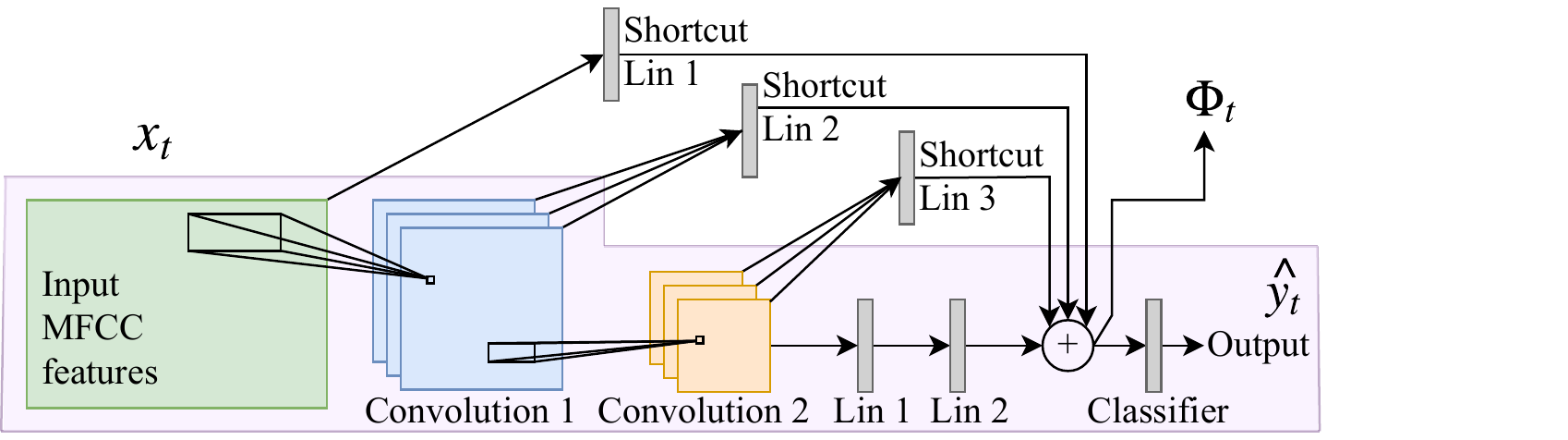}
    \caption{SANAS architecture based on \textit{cnn-trad-fpool3} \cite{DBLP:conf/interspeech/SainathP15}. Edges between layers are sampled by the model. The highlighted architecture is the base model on which we have added shortcut connections. Conv1 and Conv2 have filter sizes of (20,8) and (10,4). Both have 64 channels and Conv1 has a stride of 3 in the frequency domain. Linear 1,2 and the Classifier have 32, 128 and 12 neurons respectively. Shortcut linears all have 128 neurons to match the dimension of the classifier.}
    \label{fig:cnn_SANAS}
\end{figure}

\subsection{Stochastic Adaptive Architecture Search: Principles}

We propose now a different setting where the architecture of the model can change at each timestep depending on the context of the prediction $z_t$. At time $t$, in addition to producing a distribution over possible labels, our model also maintains a distribution over possible architectures denoted $P(\mathcal{A}_t | z_t,\theta)$. The prediction $y_t$ being now made following\footnote{$f$ is usually a distribution over possible labels.} $f(z_t,x_t,\theta,\mathcal{A}_t)$ and the context update being $z_{t+1}=g(z_t,x_t,\theta,\mathcal{A}_t)$. In that case, the cost of a prediction at time $t$ is now $C(\mathcal{A}_t)$, which also includes the computation of the architecture distribution $P(\mathcal{A}_t | z_t,\theta)$. It is important to note that, since the architecture $\mathcal{A}_t$ is chosen by the model, it has the possibility to learn to control this cost itself.  A budgeted learning problem can thus be defined as minimizing a trade-off between prediction loss and average cost. Considering a labeled sequence $(x,y)$, this trade-off is defined as :
\[
\mathcal{L}(x,y,\theta) = \mathbb{E}_{\{\mathcal{A}_t\}}\big[\sum\limits_{t=1}^{\#x} [\Delta(f(z_t,x_t,\theta,\mathcal{A}_t),y_t)+\lambda C(\mathcal{A}_t)]\big]
\]
where $\mathcal{A}_1,...,\mathcal{A}_{\#x}$ are sampled following $P(\mathcal{A}_t | z_t,\theta)$ and $\lambda$ controls the trade-off between cost and prediction efficiency. Considering that $P(\mathcal{A}_t | z_t,\theta)$ is differentiable, and following the derivation schema proposed in \cite{DBLP:journals/corr/DenoyerG14} or \cite{Veniat_2018_CVPR}, this cost can be minimized using the Monte-Carlo estimation of the gradient. Given one sample of architectures $\mathcal{A}_1,...,\mathcal{A}_{\#x}$, the gradient can be approximated by:
\begin{equation*}
\begin{aligned}
\nabla_\theta \mathcal{L}(x,y,\theta) &\approx \big(\sum\limits_{t=1}^{\#x} \nabla_\theta \log P(\mathcal{A}_t | z_t,\theta) \big) \mathcal{L}(x,y,\mathcal{A}_1,...,\mathcal{A}_{\#x},\theta)\\ 
&+\sum\limits_{t=1}^{\#x} \nabla_\theta \Delta(f(z_t,x_t,\theta,\mathcal{A}_t),y_t)
\label{eqg}
\end{aligned}
\end{equation*}
where 
\[\mathcal{L}(x,y,\mathcal{A}_1,...,\mathcal{A}_{\#x},\theta)=\sum\limits_{t=1}^{\#x} [\Delta(f(z_t,x_t,\theta,\mathcal{A}_t),y_t)+\lambda C(\mathcal{A}_t)]
\]
In practice, a variance correcting value is used in this gradient formulation to accelerate the learning as explained in \cite{DBLP:journals/ml/Williams92, DBLP:conf/icann/WierstraFPS07}.
\subsection{The SANAS Model}\label{sec:sanas}
We now instantiate the previous generic principles in a concrete model where the architecture search is cast into a sub-graph discovery in a large graph representing the search space called \textit{Super-Network} as in \cite{Veniat_2018_CVPR}.

\textbf{NAS with Super-Networks (static case):}  A Super-Network is a directed acyclic graph of layers $L=\{l_1, ... l_n\}$, of edges $E \in \{0,1\}^{n\times n}$ and where each existing edge connecting layers $i$ and $j$ ($e_{i,j}=1$) is associated with a (small) neural network $f_{i,j}$. The layer $l_1$ is the input layer while $l_n$ is the output layer. The inference of the output is made by propagating the input $x$ over the edges, and by summing, at each layer level, the values coming from incoming edges. Given a Super-Network, the architecture search can be made by defining a distribution matrix $\Gamma \in [0,1]^{n\times n}$ that can be used to sample edges in the network using a Bernoulli distribution. Indeed, let us consider a binary matrix $H$ sampled following $\Gamma$, the matrix $E \circ H$ defines a sub-graph of $E$ and corresponds to a particular neural-network architecture which size is smaller than $E$ ($\circ$ being the Hadamard product). Learning $\Gamma$ thus results in doing architecture search in the space of all the possible neural networks contained in Super-Network. At inference, the architecture with the highest probability is chosen.

\textbf{SANAS with Super-Networks:} Based on the previously described principle, our method proposes to use a RNN to generate the architecture distribution at each timestep -- see Figure \ref{fig:sanas_arch}. Concretely, at time $t$, a distribution over possible sub-graphs $\Gamma_t = h(z_t,\theta)$ is computed from the context $z_t$. This distribution is then used to sample a particular  sub-graph represented by $H_t \sim \mathcal{B}(\Gamma_t)$, $\mathcal{B}$ being a Bernoulli distribution. This particular sub-grap $E \circ H_t = \mathcal{A}_t$ corresponds to the architecture used at time $t$. Then the prediction $\hat{y_t}$ and the next state $z_{t+1}$ are computed using the functions $f(z_t,x_t,\theta, E \circ H_t)$ and $g(z_t,\Phi(x_t,\theta, E \circ H_t),\theta)$ respectively, where $g(z_t,.,\theta)$ is a classical RNN operator like a Gated Recurrent Unit\cite{DBLP:journals/corr/ChoMBB14} cell for instance and $\Phi(x_t, \theta, E \circ H_t)$ is a feature vector used to update the latent state and computed using the sampled architecture $\mathcal{A}_t$. The learning of the parameters of the proposed model relies on a gradient-descent method based on the  approximation of the gradient provided previously, which simultaneously updates the parameters $\theta$ and the conditional distribution over possible architectures.

\begin{figure}[t!]
    \centering
    \includegraphics[width=.8\linewidth]{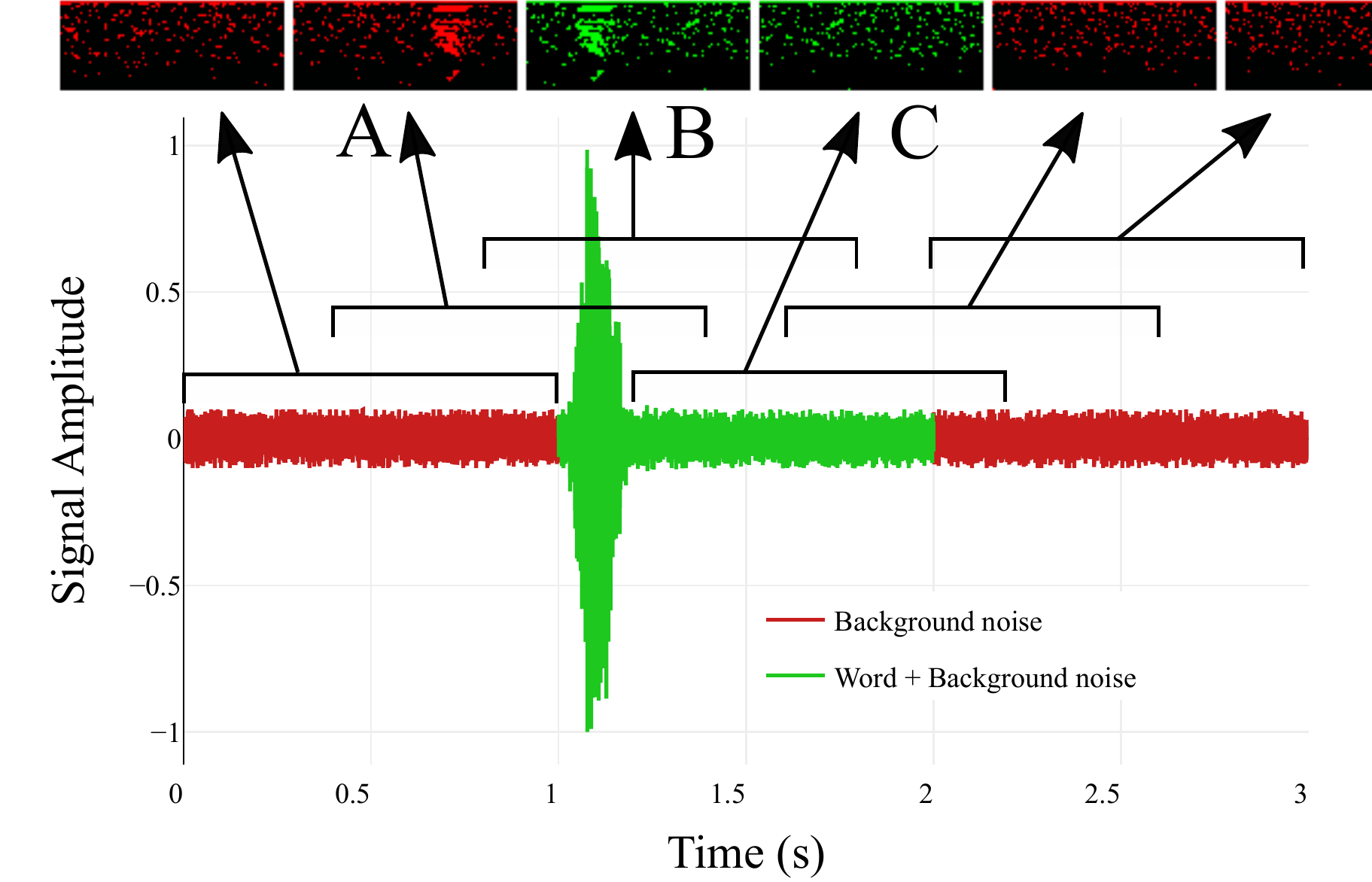} 
    \caption{Example of labeling using the method presented in section \ref{sec:experiments}. To build the dataset, a ground noise (red) is mixed with randomly located words (green). The signal is then split in 1s frames every 200ms. When a frame contains at least 50\% of a word signal, it is labeled with the corresponding word (frame B and C -- frame A is labeled as \textit{bg-noise} ). Note that this labeling could be imperfect (see frame A and C).}
    \label{fig:data}
\end{figure}
\section{Experiments}
\label{sec:experiments}

We train and evaluate our model using the Speech Commands dataset \cite{DBLP:Warden_speech_commands}. It is composed of 65000 short audio clips of 30 common words. As done in \cite{DBLP:conf/icassp/TangL18, DBLP:conf/icassp/TangWTL18, DBLP:journals/corr/HelloEdge}, we treat this problem as a classification task with 12 categories: 'yes', 'no', 'up', 'down', 'left', 'right', 'on', 'off', 'stop', 'go', '\textit{bg-noise}' for background noise and '\textit{unknown}' for the remaining words.

Instead of directly classifying 1 second samples, we use this dataset to generate between 1 and 3 second long audio files by combining a background noise coming from the dataset with a randomly located word (see Figure \ref{fig:data}), the signal-to-noise ratio being randomly sampled with a minimum of 5dB. We thus obtain a dataset of about 30,000 small files\footnote{tools for building this dataset are available at \url{http://github.com/TomVeniat/SANAS} with the open-source implementation.}, and then split this dataset in  train, validation and test sets using a 80:10:10 ratio. The sequence of frames is created by taking overlapping windows of 1 second every 200ms. The input features for each window are computed by extracting 40 mel-frequency spectral coefficients (MFCC) on 30ms frames every 10ms and stacking them to create 2D time/frequency maps. For the evaluation, we use both the prediction accuracy and the number of operations per frame (FLOPs) value. The model selection is made by training multiple models, selecting the best models on the validation set, and computing their performance on the test set. Note that the 'best models' in terms of both accuracy and FLOPs are the models located on the pareto front of the accuracy/cost validation curve as done for instance in \cite{DBLP:conf/iconip/ContardoDA16}. These models are also evaluated using the \textit{matched, correct, wrong} and \textit{false alarm} (FA) metrics as proposed in \cite{DBLP:Warden_speech_commands} and computed over the one hour stream provided with the original dataset. Note that these last metrics are computed after using a post-processing method that ensures a labeling consistency as described in the reference paper.

As baseline static model, we use a standard neural network architecture used for Keyword Spotting aka the \textit{cnn-trad-fpool3} architecture proposed in \cite{DBLP:conf/interspeech/SainathP15}  which consists in two convolutional layers followed by 3 linear layers. We then proposed a SANAS extension of this model (see Figure \ref{fig:cnn_SANAS}) with additional connections that will be adaptively activated (or not) during the audio stream processing. In the SANAS models, the recurrent layer $g$ is a one-layer GRU \cite{DBLP:journals/corr/ChoMBB14} and the function $h$ mapping from the hidden state $x_t$ to the distribution over architecture $\Gamma_t$ is a one-layer linear module followed by a sigmoid activation. The models are learned using the ADAM \cite{DBLP:journals/corr/KingmaB14} optimizer with $\beta_1=0.9$ and $\beta_2=0.999$, gradient steps between $10^{-3}$ and $10^{-5}$ and $\lambda$ in range [$10^{-(m+1)}$, $10^{-(m-1)}$] with $m$ the order of magnitude of the cost of the full model. Training time is reasonable and corresponds to about 1 day on a single GPU computer.

\begin{table}[t]
\centering
\begin{tabular}{|ccccc|}
\hline
\textbf{Match} & \textbf{Correct} & \textbf{Wrong} & \textbf{FA} & \textbf{FLOPs per frame} \\ \hline\hline
\multicolumn{5}{|c|}{\textit{\textbf{cnn-trad-fpool3}}}                           \\ \hline
81.7\%         & 72.8\%           & 8.9\%          & 0.0\%       & 124.6M       \\ \hline\hline
\multicolumn{5}{|c|}{\textbf{\textit{cnn-trad-fpool3} + shortcut connections}}                           \\ \hline
82.9\%         & 77.9\%           & 5.0\%          & 0.3\%       & 137.3M       \\ \hline\hline
\multicolumn{5}{|c|}{\textbf{SANAS}}                                              \\ \hline
61.2\%         & 53.8\%           & 7.3\%          & 0.7\%       & 519.2K       \\
62.0\%         & 57.3\%           & 4.8\%          & 0.1\%       & 22.9M        \\
86.5\%         & 80.7\%           & 5.8\%          & 0.3\%       & 37.7M        \\
86.3\%         & 80.6\%           & 5.7\%          & 0.2\%       & 48.8M        \\
81.7\%         & 76.4\%           & 5.3\%          & 0.1\%       & 94.0M        \\
81.4\%         & 76.3\%           & 5.2\%          & 0.2\%       & 105.4M       \\ \hline
\end{tabular}

\caption{Evaluation of models on 1h of audio from \cite{DBLP:Warden_speech_commands} containing words roughly every 3 seconds with different background noises. We use the label post processing and the streaming metrics proposed in \cite{DBLP:Warden_speech_commands} to avoid repeated and noisy detections. Matched \% corresponds to the portion of words detected, either correctly (Correct \%) or incorrectly (Wrong \%). FA is \textit{False Alarm}. }
\label{tab:streaming}
\end{table}

\begin{figure}[t]
    \vspace{-2.5em}
    \centering
    \includegraphics[width=.6\linewidth]{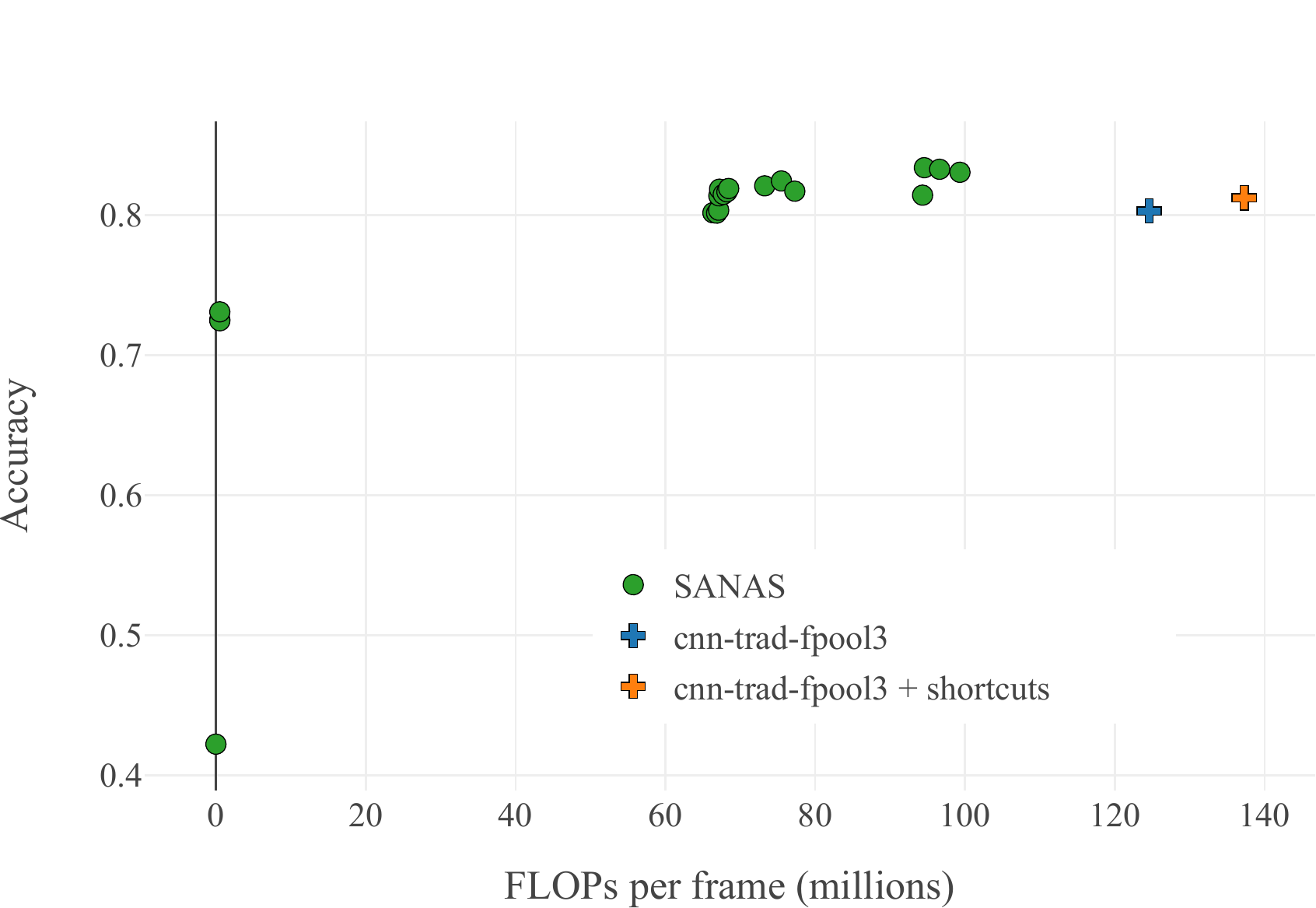}
    \caption{\textbf{Cost accuracy curves.} Reported results are computed on the test set using models selected by computing the Pareto front over the validation set. Each point represents a model.}
    \label{fig:acccost}
\end{figure}
\begin{figure}[t]
    \centering
    \includegraphics[width=.6\linewidth]{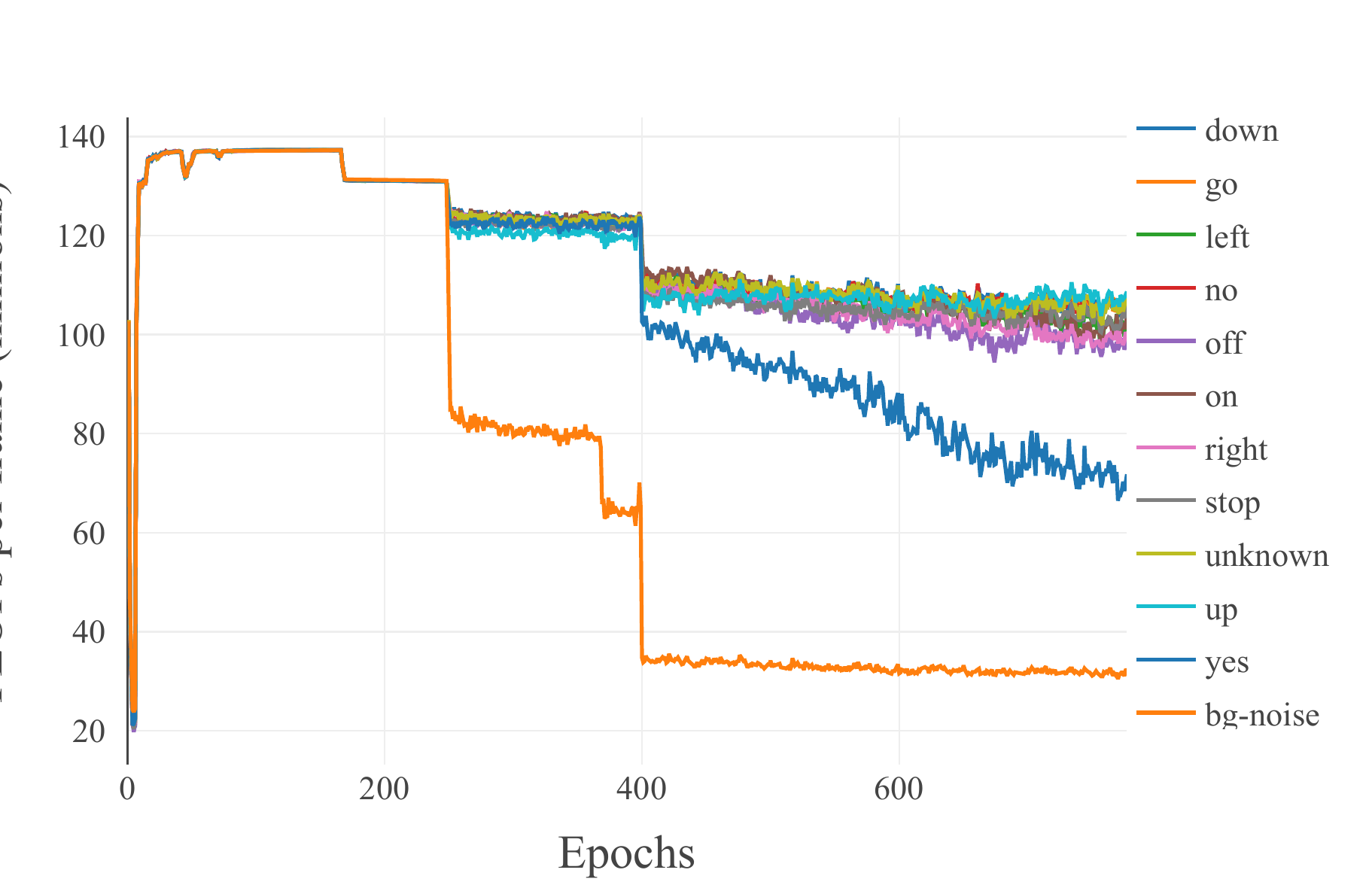} 
    \caption{\textbf{Training dynamics.} Average cost per output label during training. The network is able to find an architecture that solves the task while sampling notably cheaper architectures when only background noise is present in the frames. }
    \label{dyn}
\end{figure}




Results obtained by various models are illustrated in Table \ref{tab:streaming} for the one-hour test stream, and in Figure \ref{fig:acccost} on the test evaluation set. It can be seen that, at a given level of accuracy, the SANAS approach is able to greatly reduce the number of FLOPs, resulting in a model which is much more power efficient. For example, with an average cost of 37.7M FLOPs per frame, our model is able to match 86.5\% of the words, (80.7\% correctly and 5.8\% wrongly) while the baseline models match 81.7\% and 82.9\% of the words with 72.8\% and 77.9\% correct predictions while having a higher budget of 124.6M and 137.3 FLOPs per frame respectively.  Moreover, it is interesting to see that our model also outperforms both baselines in term of accuracy, or regarding the metrics in Table \ref{tab:streaming}. This is due to the fact that, knowing that we have added shortcut connections in the base architecture, our model has a better expressive power. Note that in our case, over-fitting is avoided by the cost minimization term in the objective function, while it occurs when using the complete architecture with shortcuts (see Figure \ref{fig:acccost}). Figure \ref{dyn} illustrates the average cost per possible prediction during training. It is not surprising to show that our model automatically 'decides' to spend less time on frames containing background noise and much more time on frames containing words. Moreover, at convergence, the model also divides its budget differently on the different words, for example spending less time on the \textit{yes} words that are easy to detect.  

\section{Conclusion}

We have proposed a new model for keyword spotting where the recurrent network is able to automatically adapt its size during inference depending on the difficulty of the prediction problem at time $t$. This model is learned end-to-end based on a trade-off between prediction efficiency and computation cost and is able to find solutions that keep a high prediction accuracy while minimizing the average computation cost per timestep. Ongoing research includes using these methods on larger super-networks and investigating other types of budgets like memory footprint or electricity consumption on connected devices.

\vfill
\pagebreak

\section*{Acknowledgments}
This work has been funded in part by grant ANR-16-CE23-0016 ``PAMELA'' and grant ANR-16-CE23-0006 ``Deep in France''.

\bibliographystyle{plainnat}
\bibliography{refs}



\end{document}